# A Convolutional Neural Network for Aspect Sentiment Classification

Yongping Xing and Chuangbai Xiao and Yifei Wu and Ziming Ding

Beijing University of Technology, Beijing, China

hsingyp@163.com, cbxiao@bjut.edu.cn
568603742@qq.com, ziming_ding@163.com

## Abstract

With the development of the Internet, natural language processing (NLP), in which sentiment analysis is an important task, became vital in information processing. Sentiment analysis includes aspect sentiment classification. Aspect sentiment can provide complete and in-depth results with increased attention on aspect-level. Different context words in a sentence influence the sentiment polarity of a sentence variably, and polarity varies based on the different aspects in a sentence. Take the sentence, "I bought a new camera. The picture quality is amazing but the battery life is too short," as an example. If the aspect is picture quality, then the expected sentiment polarity is "positive"; if the battery life aspect is considered, then the sentiment polarity should be "negative"; therefore, aspect is important to consider when we explore aspect sentiment in the sentence. Recurrent neural network (RNN) is regarded as a good model to deal with natural language processing, and RNNs has get good performance on aspect sentiment classification including Target-Dependent LSTM (TD-LSTM) ,Target-Connection LSTM (TC-LSTM) (Tang, 2015a, b), AE-LSTM, AT-LSTM, AEAT-LSTM (Wang et al., 2016).There are also extensive literatures on sentiment classification utilizing convolutional neural network, but there is no literature on aspect sentiment classification using convolutional neural network. In our paper, we develop attention-based input layers in which aspect information is considered by input layer. We then incorporate attention-based input layers into convolutional neural network (CNN) to introduce context words information. In our experiment, incorporating aspect information into CNN improves the latter's aspect sentiment classification performance without using syntactic parser or external sentiment lexicons in a benchmark dataset from Twitter but get better performance compared with other models.

## 1 Introduction

Sentiment analysis, also known as opinion mining (Nasukawa and Yi, 2003; Liu, 2012), is a basic and important task in natural language processing. Natural language processing analyzes people's opinions, sentiments, and emotions from written language.

Opinions influence almost all human activities and behaviors, and thus, sentiment analysis is applied in many business and social domains (Liu, 2012). Increasing attention is paid to sentiment analysis. Among sentiment analysis approaches, aspect sentiment analysis can provide complete and in-depth results. In this paper, we will study aspect sentiment; we will determine whether the opinions on different aspects are positive, negative, or neutral with the following sentence: "The voice quality of this phone is not good, but the battery life is long." If we consider quality, then its sentiment is negative, but when we consider battery life, then it is positive. Thus, aspect sentiment is closely related to aspect.

In this paper, when we study aspect sentiment, we not only consider the whole sentence but also look into aspect. Many studies have been conducted on aspect sentiment, some of which have used the supervised learning approach, a Supported Vector Machine based on feature engineering (Pang et al., 2002), a hierarchical classification model was also proposed (Wei and Gulla, 2010), a model based on a set of aspect dependent features for classification (Jiang et al,2011). The lexicon-based approach has also been proven to perform well (Mohammad et al., 2013). Despite the effectiveness of these approaches, aspect sentiment classification remains a challenge because it requires labor intensive features engineering or lexicon among these approaches.

In recent years, deep learning model perform well on many tasks including image processing and natural language processing. Learning word vector representations in natural language processing through neural networks language models (Bengio et al., 2003; Yih et al., 2011; Mikolov et al., 2013) also achieves huge results. Word vector representtations,1-of-V encoding (here V is the size of vocabulary) onto a lower dimensional vector space via a hidden layer, have a major influence on kinds of natural language processing tasks such as text classification (Yoon Kim,2015), speech recognition (Graves et al., 2013) and neural machine translation(Sennrich et al., 2016b).Word vector representations are feature extractors that encode semantic features of words in their dimensions in essentially. Deep learning models can deal with NLP tasks without feature processing with word vector representations. These motivate researchers to develop neural network approaches to deal with aspect sentiment classification. Aspect sentiment classification benefits from considering aspect information, such as in Target-Dependent LSTM (TD-LSTM) and Target-Connection LSTM (TC-LSTM) (Tang, 2015a, b); AE-LSTM, AT-LSTM, and AEAT-LSTM (Wang et al., 2016); AB-LSTM1 and AB-LSTM2(Yang et al.,2017).However, those models can only use RNNS for aspect sentiment classification.

In our work, we study aspect sentiment classification through convolutional neural network (CNN), which utilizes convolving filters to extract local features (LeCun et al., 1998). CNN models are proven effective for NLP and achieved excellent results in semantic parsing (Yih et al., 2014), text classification (Yoon Kim, 2015), and other traditional NLP tasks (Collobert et al., 2011); however, no study on aspect classification has used convolutional neural network. Previous CNN models have treated input sentences but ignored aspect information. We solve this problem through an input layer attention mechanism (Bahdanau et al., 2014). Attention mechanism is effective for obtaining good results, as demonstrated in machine translation (Bahdanau et al., 2014), entailment reasoning (Rocktaschel et al., 2015) and sentence summarization (Rush et al., 2015), so we incorporate an attention-based input layer into CNN to improve aspect classification performances, and the experiment results show that our model achieves state-of-the-art results in a benchmark dataset from Twitter.

Our paper is structured as follows: Section 2 discusses related works, Section 3 introduces our CNN model in detail, Section 4 implements experiments to test and verify the effectiveness of our model, and discusses the detail of experiments, and Section 5 summarizes our work .

## 2 Related work

Different studies on aspect sentiment classification will be discussed in this section.

Aspect sentiment classification is a classic text classification problem in natural language processing(NLP), which can be solved by text classification approaches. Current models try to judge the polarity of the sentence ignoring the aspect mostly. Lots of models deal with the problem by a set of features; among them are Supported Vector Machine (Pang et al., 2002; Jiang et al., 2011), which is effective but requires labor intensive feature engineering, and sentiment analysis, which is based on the lexicon features (Mohammad et al., 2013) with sentiment lexicons (Rao et al., 2009; Kaji et al.,2007).All these approaches require labor intensive feature engineering, and thus, some other approaches are motivated to deal with the problems. In recent years, deep learning model perform well on many tasks including image processing and natural language processing. Learning word vector representations, which play an important role in natural language processing, through neural networks language models also achieves huge results. Neural networks advanced sentiment analysis in various fields, Recursive Neural Network (Socher et al., 2011; Donget et al., 2014; Vo et al., 2015) encodes each sentence in low-dimensional vector space without feature engineering and then gets the better sentence representation to classify the sentence. In addition, Vo and Zhang (2015) also use features including sentiment-specific word embedding and sentiment lexicons.

In recent literatures some neural models are studied in order to avoid labor intensive feature engineering, RNNs can solve the serial problem so they are regarded as one of the best approaches to deal with NLP tasks. Long Short-Term Memory abbreviated to "LSTM" (Hochreiter and Schmidhuber, 1997), and its derivatives are used to solve the problem. Target-Dependent LSTM (TD-LSTM) and Target-Connection LSTM (TC-LSTM) (Tang, 2015a, b) considered target information and got good performance. Tree-LSTMs (Tai et al., 2015) utilizes the syntax structures of sentences but introduces syntax parsing errors. LSTM models can be purely data-driven and do not rely on dependency parsing results, external sentiment lexicons, or labor intensive feature engineering. In addition, CNNs also are utilized in classifying text (Kalchbrenneret al., 2014; Kim, 2014), which can model N-gram language model with varying filter lengths and then get the most important feature through Max-poolings. But there are not CNN models solve the aspect sentiment classification considering aspect, so we are motivated to design a powerful CNN model which can get better performance with other models.

## 3 Model

In this section, we will talk our model in details.

### 3.1 N-gram language model and convolutional neural networks

Language model research is a very important task in natural language processing, which plays a highly important pole in many NLP tasks, such as language machine translation, automatic speech recognition, handwriting recognition, text classification and information retrieval. Language model includes natural language model based on rules and statistical language model. Statistical Language model is the probability distribution of a word sequences; which can estimate probability on word sequences that will occur in automatic sentence generation. Statistical Language model has outstanding performance. The performance of a language model can be evaluated by the empirical perplexity. The goal of language model is to obtain a small perplexity which means perplexity is better when the perplexity is smaller. There are kinds of language models in literatures, but the effective models are few, one of the simplest and most successful language model is the N-gram language model. Given a word sequence $w_1, w_2 \dots w_n$, n is the length of the sequence, note that by the chain rule of probability we can write the probability of any sequence as

$$P(w_1 w_2 \dots w_n) = \prod_{i=1}^{n} P(w_i | w_1 w_2 \dots w_{i-1}) \tag{1}$$

the perplexity is

$$\text{Perplexity} = \sqrt[n]{\prod_i^n \frac{1}{P(w_1 w_2 \dots w_n)}} \tag{2}$$

N-gram model approximates probability of the ith word $w_i$ $P(w_1 w_2 \dots w_n)$ by assuming that $P(w_1 w_2 \dots w_n)$ can be approximated by the probability of observing it in the shortened context history of the preceding n − 1 words. This is to say, the relevant words of $w_i$ are the just previous n − 1 words in the context history,

$$P(w_1 w_2 \dots w_n) = P(w_i | w_1 w_2 \dots w_{i-1}) = P(w_i | w_{i-n+1} \dots w_{i-1}) \tag{3}$$

n can be 1,2,3, …,which can be called unigram, bigram, trigram,… In general, N is less than be 5. N-gram language model play an important role in natural language processing.

Convolutional neural network (CNN, or ConvNet) is a class of deep, feed-forward artificial neural networks which has successfully been applied to analyzing visual imagery and image processing. CNN use a variation of multilayer perceptron designed to require minimal preprocessing (LeCun,Yann,2013), They are also known as shift invariant or space invariant artificial neural networks (SIANN), based on their shared-weights architecture and translation invariance characteristics(Zhang et al.1988; Zhang et al.1990).CNN can solve visual imagery and image processing with little pre-processing which means independence on prior knowledge is its major advantage.

In recent years, CNN is also used in NLP its advantage. Convolutional neural network can extract language features at different positions by a filter vector sliding over a sequence, the length of filter vector size can be regarded as N in N-gram language model, so we can detect language feature information through convolution neural network. Let x∈ $R^{L*d}$ denote the input sentence where L is the length of the sentence, and $x_i \in R^d$ be the d-dimensional word vectors for the i-th word in the input sentence. Let k be the length of the filter, and the vector m∈ $R^{k*d}$ is a filter for convolution operation. For each position j in the sentence, we have a window vector $w_j$ with k consecutive word vectors, denoted as:

$$w_j = [x_j, x_{j+1}, \ldots, x_{j-k+1,}] \tag{4}$$

The commas represents row vector concatenation. A filter m convolves with the window vectors (k-grams) at each position in a valid way to generate a feature map c∈ $R^{l-k+1}$, each element $c_j$ of the feature map c for window vector $w_j$ is produced as follows:

$$c_j = f(w_j \circ m + b) \tag{5}$$

where ○ is element-wise multiplication, b is a bias term and f is a nonlinear transformation function, called activation function in neural networks, that can be sigmoid, TanH ,hyperbolic tangent, ArcTan, etc. In our model, we choose ReLU (Nair and Hinton, 2010) as activation function. The length and number of filter can vary in convolutional neural network. For n filters with the same length, their n feature maps can be rearranged as:

$$W = [c_1; c_2; \ldots c_n] \tag{6}$$

$c_i$, is the feature map generated with the i-th filter. Generally, a pooling layer often applied to feature maps after the convolution to capture features, max-over-poolings or dynamic k-max poolings are usually used to select the most or the k-most important features. These features are passed to a fully connected softmax layer whose output is the probability distribution over labels.

In our model, we use multiple window lengths in order to capture features information at different positions to implement n-gram language through convolutional neural networks. The window size means how many words are matched by the filter, and then, we adopt max pooling as pooling layer over the convolutional results. Max pooling make convolutional model capture the most prominent and prevalent features, which is very useful for convolutional model to keep its robustness.

## 3.2 Attention-based input layer

Attention is the behavioral and cognitive process of selectively concentrating on a discrete aspect of information, whether deemed subjective or objective, while ignoring other perceivable information. It is the taking possession by the mind in clear and vivid form of one out of what seem several simultaneous objects or trains of thought. Focalization, concentration of consciousness are of its essence. Attention mechanism allocates limited processing resources upon concentrating on a discrete aspect of information. Attention mechanism paly an important role in kinds of fields including information processing. Attention mechanism has obtained good results in many cases, which demonstrated in image recognition (Mnih et al., 2014), reasoning on entailment (Rockt¨aschel et al., 2015), machine translation (Bahdanau et al., 2014), sentence summarization (Rush et al., 2015), read comprehension (Hermann et al., 2015).

In our paper, we will introduce an attention mechanism and our attention mechanism is based on input layer that converts word of input sentences into an aspect-related word to enforce the model to attend to the importance of aspect of the sentence. The input layer attention mechanism can concentrate on the key part of a sentence given the aspect, which allocates limited processing resources upon concentrating on a discrete aspect of information .To be more specific,It is applied over word embeddings of input sentences to generate re-weighted word embeddings. The attention mechanism can focus on important input words by re-weighting word embeddings of the input sentence. That is, words in one sentence that are more relevant to aspect receive more higher weights.

Let $x_i \in R^d$ be the d-dimensional word embeddings for the i-th word in a sentence. Let $x \in R^{L*d}$ denotes the input sentence where L is the length of the sentence, $a \in R^d$ denotes d-dimensional word embeddings for the aspect in the sentence, when there are more than one aspect word, aspect word embeddings is the mean of the all aspect word embeddings. We then define an attention $D=R^L$, $D_i$ represents the word similarity score between the i-th word embeddings of x and the aspect word embeddings a . The similarity score uses cosine distance:

$$D_i = \text{cosine}(x_i, a) \qquad (7)$$

$A_i$ can be regarded as as an attention-based relevance score of one word embedding $x_i$ with respect to aspect embedding of the sentence. We utilize the softmax normalization on attention weights:

$$E = (D_1, D_2, \dots D_L) \qquad (8)$$
$$A_i = \text{softmax}(E) \qquad (9)$$

Then we updated embeddings attenEmb $\in R^{2d}$ for every word by a concatenation of the original word embeddings and attention-reweighted word embeddings,

$$\text{attenEmb}_1 = \text{contact}(x_i; A_i \odot x_i) \qquad (10)$$

where $\odot$ represents element-wise multiplication.

We introduce attention mechanism by input layer simply using cosine distance to create the attention weights simply using cosine distance to create the attention weights and generate attenEmb, it is proved to be simple but effective. Moreover, we do not introduce any additional parameters. There is another manner to concentrate on some parts of information by introducing relevance from aspect to each word of the sentence:

$$\text{attenEmb}_2 = \text{contact}(x_i; a) \qquad (11)$$

$\text{attenEmb}_2$ is even more simpler but more effective in performance. In this way aspect words are paid more attention to deal with aspect sentiment classification, it is proved effective in lstm on sentiment chassification and it is plausible.

Based discussion above, we give our model architecture in the following figure:

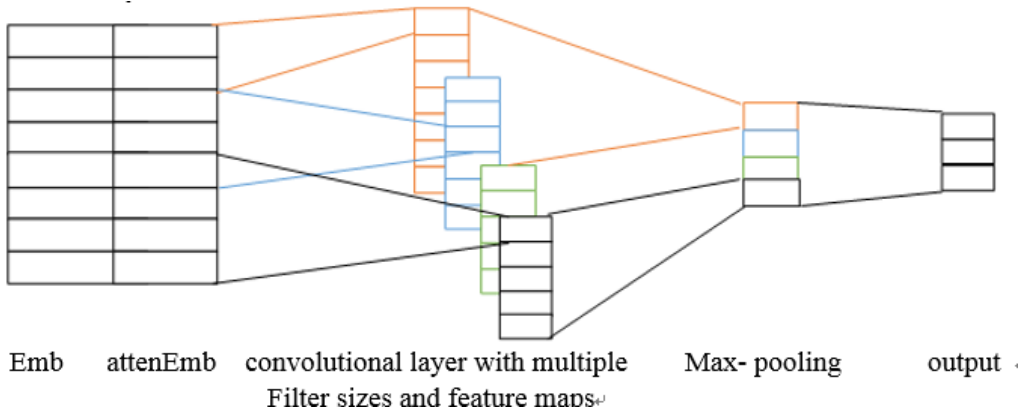

Figure 1: Illustration of a CNN architecture for sentence classification. Input sentence convolutes with multiple filter sizes, and then imply Max-pooling on the results to get the sentence representation, which is used to classify the sentence.

## 2.3 Model training

In our model, we add a fully connected softmax layer, whose output is the probability distribution, on top of the sentence features to predict sentiment distribution of the sentence. We train the model by minimizing loss function (Rosasco, Lorenzo, et al.2004) where we use cross-entropy loss as loss function. Given a training sample x, let y be the ground truth distribution for sentence, $\hat{y}$ be the predicted sentiment distribution. The goal of model training is to minimize the cross-entropy loss between y and $\hat{y}$ for all sentences.

$$\text{loss} = \sum_i \sum_j y_i^j \log(\hat{y}_i^j) + \lambda ||\theta||^2 \tag{12}$$

where j is the index of class, which is positive or neutral or negative, i is the index of sentence, $\lambda$ is the $L_2$-regularization term, $\theta$ is the parameter set. We impose dropout, which can prevent over-fitting by randomly dropping out, on the penultimate layer with a constraint on L2-norms of the weight vectors (Hinton et al., 2012). Dropout randomly sets hidden units to 0 during forward-back propagation with a probability of p.

## 4 Experiment and results

In this section, will talk about experiment details.

### 4.1 Dataset

In this paper, we implement our experiment on an original dataset collected from Twitter (Dong et al., 2014) where every instance has been labeled sentiment polarity manually, to evaluate the performance of our approach. Our aim is to identify the aspect polarity of a sentence with the corresponding aspect. The dataset include training set and test set, training set contains 6,248 sentences and test set contains 692 sentences. The percentages of positive, neutral and negative in training data and test data are both 25%, 50%, 25%. We train the model on training set, and then compute the performance of the model on test set. Evaluation metrics of our model are accuracy and macro-F1 score over positive, neutral and negative categories (Jurafsky and Martin, 2000; Manning and Schutze, 1999). We preprocess the data by removing non-alphabet characters, punctuation, numbers, and stop words from the sentence for the dataset.

### 4.2 Experimental Task Definition

In our papers, we define aspect sentiment classification task as determining the polarity of each aspect term for a given set of aspects term within a sentence. For example, the sentiment polarity of aspect term picture quality is positive in the sentence "I bought a new camera. The picture quality is amazing but the battery life is too short".

### 4.3 Experimental Settings

In our experiments, we initialize all word vectors by Glove (Pennington et al.,2014), where the word embedding vectors are pre-trained on an unlabeled twitter corpus. We set dimension of word embedding vectors and aspect word embedding vectors to be 200. Words not present in the set of pre-trained words are initialized randomly. We initialize other parameters by sampling from a uniform distribution .We set maxlen as the maximum length of the sentence in the training set, if the sentence in the training dataset has a length less than maxlen we pad the sentence with special symbols at the end which indicate the unknown words, we also pad sentences that are shorter than maxlen for a sentence in the test dataset in the same way, and we simply cut extra words at the end of these sentences to reach maxlen for sentences in the training set and test set, this is because the convolution layer requires fixed-length input in our model. In our experiments, we used tensorflow for implementing our models. We use multiple convolutional layers with different lengths of filters in parallel, where the filter length is (2, 3, 4), and the number of filters is set to be 200. The dropout probability of the penultimate layer is 0.5 when we train model, and it is set to be 1 when test performance of our model. We implement L2-regularization on the softmax layer with weight of 2.6. We train our models with a batch size of 64 examples, and use Adam as our optimization method, which has improved the robustness of SGD on large scale learning task remarkably, and initialize the learning rate by 0.001.

### 4.4 Experimental results

In our paper, we regard SVM (Pang, Lee, and Vaithyanathan, 2002), SVM with target-dependent features (SVM-dep) (Jiang et al., 2011), TC-LSTM, and TD-LSTM (Tang et al. 2015) as baseline methods to compare with our approach. SVM classifier is built with many features, such as n-gram, punctuations, hashtags, and the numbers of positive or negative words in sentiment lexicon. TD-LSTM splits the sentence by the target term of the sentence and contacts the representation of preceding part and the following part of the aspect term as the whole sentence representation, which means it cannot pay attention to which words are more important for the given aspect term.TC-LSTM improve the performance of TD-LSTM by incorporating aspect term information into the representation of a sentence by adding target word vectors obtained from word vectors into the LSTM cell unit input. We set the mean of all word vectors as aspect term vector when the aspect term has more than one word.

Experimental results of baseline models and our methods are given in the following Table.

| Model | Accruacy | Mcro-F1 |
|---|---|---|
| SVM | 0.627 | 0.602 |
| SVM-dep | 0.634 | 0.633 |
| TD-LSTM | 0.708 | 0.690 |
| TC-LSTM | 0.715 | 0.695 |
| attenEmb$_1$ | 0.716 | 0.700 |
| attenEmb$_2$ | 0.725 | 0.702 |

Table 1: Comparison of different methods on aspect sentiment classification. Accuracy and macro-F1 are evaluation metrics.

From table 1, we can conclude that our approaches achieve better results on the experimental dataset compared to previous methods.

## 5 Conclusion and discussion

We develop a model which incorporates attention-based input layer into convolutional neural networks to improve the performance of aspect sentiment classification. Experiments on dataset from twitter showed that our model achieves better accruacy and mcro-F1.

**Reference**

Tetsuya Nasukawa and Jeonghee Yi. 2003. Sentiment analysis: Capturing favorability using natural language processing. In Proceedings of the 2$^{nd}$ international conference on Knowledge capture, pages 70–77. ACM.

Bing Liu. 2012. Sentiment analysis and opinion mining. Synthesis Lectures on Human Language Technologies, 5(1):1–167.

Wei, Wei and Jon Atle Gulla. Sentiment learning on product reviews via sentiment ontology tree. In Proceedings of annual meeting of the association for computational Linguistics (acL-2010). 2010.

Jiang, Long, Mo Yu, Ming Zhou, Xiaohua Liu, and Tiejun Zhao. Target-dependent twitter sentiment classification. In Proceedings of the 49th annual meeting of the association for computational Linguistics (acL-2011). 2011.

Rico Sennrich, Barry Haddow, and Alexandra Birch. Neural machine translation of rare words with subword units. In ACL. 2016b

Duyu Tang, Bing Qin, Xiaocheng Feng, and TingLiu. 2015a. Target-dependent sentiment classification with long short term memory. arXiv preprintarXiv :1512.01100.

Duyu Tang, Bing Qin, and Ting Liu. 2015b. Document modeling with gated recurrent neural network for sentiment classification. In Proceedings of the 2015 Conference on Empirical Methods in Natural Language Processing, pages 1422–1432.

Yequan Wang , Minlie Huang , xiaoyan zhu . Attention-based LSTM for Aspect-level Sentiment Classification. Conference on Empirical Methods in Natural Language

Processing, pages 606-615

Min Yang, Wenting Tu, Jingxuan Wang, Fei Xu, Xiaojun Chen. Attention-Based LSTM for Target-Dependent Sentiment Classification. Proceedings of the Thirty-First AAAI Conference on Artificial Intelligence (AAAI-17), pages 5013-5014

Y. Bengio, R. Ducharme, P. Vincent. 2003. Neural Probabilitistic Language Model. Journal of Machine Learning Research 3:1137–1155.

W. Yih, K. Toutanova, J. Platt, C. Meek. 2011. Learning Discriminative Projections for Text Similarity Measures. Proceedings of the Fifteenth Conference on Computational Natural Language Learning, 247–256.

T. Mikolov, I. Sutskever, K. Chen, G. Corrado, J. Dean. 2013. Distributed Representations of Words and Phrases and their Compositionality. In Proceedings of NIPS 2013.

Y. LeCun, L. Bottou, Y. Bengio, P. Haffner. 1998. Gradient-based learning applied to document recognition. In Proceedings of the IEEE, 86(11):2278–2324, November.

W. Yih, X. He, C. Meek. 2014. Semantic Parsing for Single-Relation Question Answering. In Proceedings of ACL 2014.

Yoon Kim. 2014. Convolutional neural networks for sentence classification. In Proceedings of Empirical Methods on Natural Language Processing.

Ronan Collobert, Jason Weston, Léon Bottou, Michael Karlen, Koray Kavukcuoglu, and Pavel Kuksa. 2011. Natural language processing (almost) from scratch. The Journal of Machine Learning Research, 12:2493–2537.

Nobuhiro Kaji and Masaru Kitsuregawa. 2007. Building lexicon for sentiment analysis from massive collection of html documents. In EMNLP-CoNLL, pages 1075–1083.

Delip Rao and Deepak Ravichandran. 2009. Semisupervised polarity lexicon induction. In Proceedings of the 12th Conference of the European Chapter of the Association for Computational Linguistics, Association for Computational Linguistics, pages 675–682.

Duy-Tin Vo and Yue Zhang. 2015. Target-dependent twitter sentiment classification with rich automatic features. IJCAI

Hua He, Kevin Gimpel, and Jimmy Lin. 2015. Multiperspective sentence similarity modeling with convolutional neural networks. In Proceedings of the 2015 Conference on Empirical Methods in Natural Language Processing, pages 1576–1586.

Dzmitry Bahdanau, Kyunghyun Cho, and Yoshua Bengio. 2014. Neural machine translation by jointly learning to align and translate. CoRR, abs/1409.0473.

Tim Rocktaschel, Edward Grefenstette, Karl Moritz Hermann, Tomás Kočiský, and Phil Blunsom. 2016. Reasoning about entailment with neural attention. In Proceedings of the 4th International Conference on Learning Representations.

Alexander M. Rush, Sumit Chopra, and Jason Weston. 2015. A neural attention model for abstractive sentence summarization. In Proceedings of the 2015 Conference on Empirical Methods in Natural Language Processing, pages 379–389.

Vinod Nair and Geoffrey E Hinton. 2010. Rectified linear units improve restricted boltzmann machines. In Proceedings of the 27th International Conference on Machine Learning (ICML-10), pages 807–814.

Karl Moritz Hermann, Tomas Kocisky, Edward Grefenstette, Lasse Espeholt, Will Kay, Mustafa Suleyman, and Phil Blunsom. 2015. Teaching machines to read and comprehend. In Advances in Neural Information Processing Systems, pages 1684–1692.

Li Dong, Furu Wei, Chuanqi Tan, Duyu Tang, Ming Zhou, and Ke Xu. 2014. Adaptive recursive neural network for target-dependent twitter sentiment classification. In ACL, pages 49–54.

Christopher D Manning and Hinrich Sch ̈utze. 1999. Foundations of statistical natural language processing. MIT press.

Dan Jurafsky and James H Martin. 2000. Speech & language processing. Pearson Education India.

Jeffrey Pennington, Richard Socher, and Christopher D Manning. 2014. Glove: Global vectors for word representation. In EMNLP, pages 1532–1543.

Bo Pang, Lillian Lee, and Shivakumar Vaithyanathan. 2002. Thumbs up?: sentiment classification using machine learning techniques. In EMNLP, pages 79–86.

Long Jiang, Mo Yu, Ming Zhou, Xiaohua Liu, and Tiejun Zhao. 2011. Target-dependent twitter sentiment classification. ACL, 1:151–160.

LeCun, Yann.2013.LeNet-5, convolutional neural networks".

Zhang, Wei .1988. "Shift-invariant pattern recognition neural network and its optical architecture". Proceedings of annual conference of the Japan Society of Applied Physics.

Zhang, Wei .1990. Parallel distributed processing model with local space-invariant interconnections and its optical architecture. Applied Optics. 29 (32): 4790–7. Bibcode:1990 ApOpt..29.4790Z. doi:10.1364/AO.29.004790. PMID 20577468.

Rosasco, Lorenzo, et al.2004. Are loss functions all the same? Neural Computation 16.5: 1063-1076.

G. Hinton, N. Srivastava, A. Krizhevsky, I. Sutskever,R. Salakhutdinov. 2012. Improving neural networks by preventing co-adaptation of feature detectors. CoRR, abs/1207.0580.